\documentclass{ifacconf}

\usepackage{graphics} 
\usepackage{graphicx}
\usepackage{natbib}        
\usepackage{amsfonts}
\usepackage{amsmath}
\usepackage{epsfig} 
\usepackage{amssymb}  
 \usepackage{algorithm}
\usepackage{algpseudocode}
\algnewcommand\AAND{\textbf{ and }}
\algnewcommand\Or{\textbf{ or }}
\usepackage{color}
\usepackage{flushend}
\usepackage{url}

\usepackage[table,xcdraw]{xcolor}

\DeclareMathAlphabet{\pazocal}{OMS}{zplm}{m}{n}

\newcommand{\Xs}{\pazocal{X}}

\newcommand{\Ms}{\pazocal{M}}

\newcommand{\BigO}{\mathcal{O}}
\newtheorem{definition}{Definition}

\usepackage{array}
\newcolumntype{C}[1]{>{\centering\arraybackslash}p{#1}}
\newcolumntype{M}[1]{>{\raggedright\arraybackslash}p{#1}}

\usepackage{array} 
\newcolumntype{L}[1]{>{\raggedright\let\newline\\\arraybackslash\hspace{0pt}}m{#1}}	
\newcolumntype{S}[1]{>{\centering\let\newline\\\arraybackslash\hspace{0pt}}m{#1}}
\newcolumntype{R}[1]{>{\raggedleft\let\newline\\\arraybackslash\hspace{0pt}}m{#1}}

\algnewcommand\pushup{\vspace{-1ex}}
\algnewcommand\pushuphalf{\vspace{-0.5ex}}

\newtheorem{problem}{\textbf{Problem}}

\makeatletter

\DeclareMathAlphabet{\mathpzc}{OT1}{pzc}{m}{it}
\begin{document}
\begin{frontmatter}

\title{The Reconfigurable Aerial Robotic Chain: Shape and Motion Planning} 

\thanks[footnoteinfo]{This material is based upon work related to the Mine Inspection Robotics project sponsored by the Nevada Knowledge Fund administered by the Governor's Office of Economic Development.}

\author[First]{Mihir Kulkarni}  
\author[Second]{Huan Nguyen} 
\author[Third]{Kostas Alexis}

\address[First]{Birla Institute of Technology and Science (BITS) Pilani, Goa Campus, India, (e-mail: f20160150@goa.bits-pilani.ac.in).}
\address[Second]{Autonomous Robots Lab, University of Nevada, Reno, USA, (e-mail: huann@nevada.unr.edu)}
\address[Third]{Autonomous Robots Lab, University of Nevada, Reno, USA, (e-mail: kalexis@unr.edu)}

\begin{abstract}                
This paper presents the design concept, modeling and motion planning solution for the aerial robotic chain. This design represents a configurable robotic system of systems, consisting of multi-linked micro aerial vehicles that simultaneously presents the ability to cross narrow sections, morph its shape, ferry significant payloads, offer the potential of distributed sensing and processing, and allow system extendability. We contribute an approach to address the motion planning problem of such a connected robotic system of systems, making full use of its reconfigurable nature, to find collision free paths in a fast manner despite the increased number of degrees of freedom. The presented approach exploits a library of aerial robotic chain configurations, optimized either for cross-section size or sensor coverage, alongside a probabilistic strategy to sample random shape configurations that may be needed to facilitate continued collision-free navigation. Evaluation studies in simulation involve traversal of constrained and obstacle-laden environments, having narrow corridors and cross sections.
\end{abstract}


\end{frontmatter}

\section{INTRODUCTION}\label{sec:intro}
\begin{figure}[h!]
\centering
    \includegraphics[width=0.95\columnwidth]{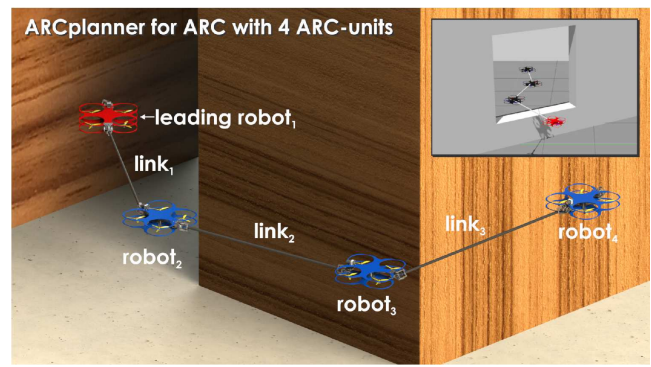}
\caption{Visualization of the aerial robotic chain with 4 ARC-units. A video of the conducted simulations is available at \protect\url{https://youtu.be/wP8Mg9_YEh8}}\label{fig:arcplannerintroduction}
\end{figure}

Research in aerial robotics has enabled their widespread utilization in a multitude of applications including infrastructure inspection~\cite{SIP_AURO_2015}, exploration~\cite{GBPLANNER_IROS_2019}, surveillance~\cite{grocholsky2006cooperative}, entertainment~\cite{alonso2012object} and more. One new trend in the field is to develop flexible, reactive platforms that can adapt to different tasks or environments by morphing their shapes. Previous relevant work has demonstrated quadrotor Micro Aerial Vehicles (MAVs) that can fold their arms to pass through narrow windows~\cite{falanga2018foldable,bucki2019rapid}, lattice-based~\cite{oung2014distributed} and chain-based~\cite{zhao2018design} modular platforms that can reconfigure their shapes in response to a given task. As compared to single MAV platforms, multi-system modular designs present the advantages that they can integrate significantly more payload, distribute their sensing and processing capabilities and thus also facilitate redundancy. Furthermore, chained multi-linked MAV platforms have the ability to traverse through narrow sections more effectively than either monolithic larger vehicles or lattice-based platforms. Motivated by the aforementioned benefits, in this work we investigate the problem of motion planning for a new type of reconfigurable chain-like multi-linked aerial robotic system-of-systems, called the Aerial Robotic Chain (ARC) depicted in Figure~\ref{fig:arcplannerintroduction}. While most of the work in the domain of reconfigurable flying robots has focused only on the problems of modeling and control, in this paper we deal with the complexity of the path planning given the multiple Degrees-of-Freedom (DOF) involved. 

%
%

The design of ARC consists of individual quadrotor MAVs (``ARC-units'') connected through rigid links and 3-DoF joints. Each ARC-unit is considered to integrate the necessary low-level autopilot functionalities (attitude and thrust control), alongside processing and sensing payloads. These ARC-units are then connected with each other through rigid links. The proposed motion planning approach (ARCplanner) identifies navigation solutions for the Aerial Robotic Chain with an arbitrary number of units through a bifurcated approach of a) utilizing a library of fixed configurations, generated either to minimize cross-section or to maximize sensor coverage given a certain frustum model for the perception system onboard each ARC-unit, as well as b) a random sampling-based search algorithm responsible for identifying collision-free paths when such solutions are not possible on the basis of the library-stored configurations. Given the presented abstraction of shape controller, the planning method discussed in this work can be extended to other reconfigurable robotic systems. 

In further detail, one of the main challenges in path planning for such a multi-linked robotic system-of-systems is the underlying high dimensionality of the configuration space especially for a large number of ARC-units. The proposed solution greatly addresses this challenge and the associated curse of dimensionality. First, we decouple the translation planner from the configuration (shape) planner. Second, for each path segment of the translation planning solution, we first check in the library of shape configurations if a stored shape solution exists that can be reached and traversed through that segment in a collision-free manner. Third, only if such a solution cannot be identified, we invoke a search within a set of random shape configurations sampled a priori (offline) through a Probabilistic RoadMap (PRM) based configuration planner that identifies novel shape configurations for ARC to faciliate collision-free navigation through the specified segment. Finally, if and only if all options of the local planner fail, we remove the specific segment from the solution and re-run the translation planner to identify new topologies in which possible shapes of ARC can allow collision-free flight. This approach has several advantages, namely a) on average, it is orders of magntitude faster than traditional full-state PRM since we do opportunistic search in the library of shape configurations first, and only if needed, further search within the a priori stored set of random shapes, and b) in the worst case, the translation planner is invoked again to provide new candidate topologies to be examined for allowing full collision-free navigation of ARC. 

The remainder of this paper is organized as follows: Section~\ref{sec:literature} overviews related work, while Section~\ref{sec:ProbStat} defines the motion planning problem. Section~\ref{sec:modelling} describes the model and control for the ARC system, followed by the description of proposed approach in Section~\ref{sec:approach}. Finally, extensive simulation studies are presented in Section~\ref{sec:eval}, while conclusions are drawn in Section~\ref{sec:concl}.

\section{RELATED WORK}\label{sec:literature}
ARC involves multiple degrees of freedom hence full sampling of the configuration space online is challenging~\cite{choset2005principles}. Our robot can be approximated as a deformable object for which the dimensionality of the configuration space can be reduced. \cite{DeformablePCA} uses principal component analysis (PCA) to reduce the dimension of deformation space. However, PCA can add narrow passages to the problem and create implausible deformations of the reconstructed robot.~\cite{physicsbased} assumes certain joint angles to be fixed, while simulating forward dynamics which may not be valid in our case.~\cite{DeformableKavraki} proposes an efficient representation for deformable linear objects and a path planner for them by reusing a local shape planner in a global roadmap. The community of reconfigurable robots has introduced several path planning methods.~\cite{howie99leaderfollower} uses a leader-follower approach in which the path of the first robot (the head) is computed from a Generalized Voronoi Graph (GVG).~\cite{dragon2018planner} proposes a path planner based on differential kinematics but the assumption that obstacles can be represented as primitive geometries may not hold in general.~\cite{bhat2006hierarchical} reduces the task of searching the whole configuration space by implementing a base planner for a few modules and reuse it within a hierarchical planner, however this method is only applicable to reconfigurable robots having self-similar structure. Our approach is closest to~\cite{Daudelineaat2018integrated} which uses a high-level planner to determine which of its defined configurations is best for on a current task or environment, combined with a local configuration planner. However, our approach is different in that a) we focus on cluttered environments, hence if the configuration library fails to find a solution, we utilize the solutions of a local PRM-based configuration planner to find a valid shape, and b) we replan the translation path when the local planner does not find a solution. 

\section{PROBLEM STATEMENT}\label{sec:ProbStat}
This work aims to enable collision-free motion planning for the ARC multi-linked connected system-of-systems of flying robots. The following environment representation and problem definition are considered. 

\begin{definition}[Environment Representation]\label{def:occupancymap}
Given an environment of finite volume $V_{M}$, it is assumed to be represented through a known occupancy map $\Ms$ based on a discretization of voxels with edge length $r_v$. The map is organized in a subset of obstacle space leading to occupied voxels $\Ms_{occ}$, alongside collision-free space $\Ms_{free}$. 
\end{definition}

\begin{problem}[ARC Navigation]\label{prob:arcmplanning}
Considering a dynamic model representation for the Aerial Robotic Chain system-of-systems $\dot{x}=f(x,u)$, where $x$ is the state and $u$ is the control, $x_0$ the current robot configuration, and given the obstacle space $\Ms_{occ}$ and the goal region $\Ms_{goal}$ then we have the associated space of in-collision configures $\Xs_{obs}$ and the goal configurations $\Xs_{goal}$. Given the above, the objective is to find, if it exists, a sequence of configurations $\{x\}$ leading to a path $\sigma$ such that the solution satisfies $\sigma \notin \Xs_{obs}$ and the final configuration $x_f$ of the path $\sigma$ is such that $x_f \in \Xs_{goal}$. 
\end{problem}

\section{MODELLING AND CONTROL}\label{sec:modelling}
\begin{figure}[h!]
\centering
    \includegraphics[width=0.98\columnwidth]{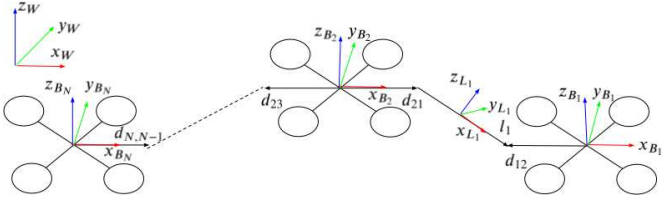}
\caption{Model diagram of the aerial robotic chain.}\label{fig:modeldiagram}
\vspace{-1ex}
\end{figure}

Let $\mathbb{W}$ denote the world frame, $\mathbb{B}_i$ the body-fixed frame of ARC-unit $i$, and $\mathbb{B}_{L_i}$ be the body-fixed frame of the connecting link $i$, as shown in Figure~\ref{fig:modeldiagram}. 

%
%

\begin{figure}[h!]
\centering
    \includegraphics[width=0.99\columnwidth]{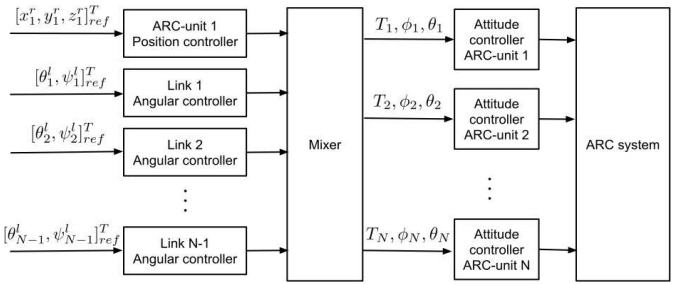}
\caption{Control diagram for the Aerial Robotic Chain. An MPC position controller and $N-1$ $\mathpzc{SO}(3)$ angular controllers receive reference inputs from the planner, the output of these controllers are mixed together to generate thrust ($T_i$), roll ($\phi_i$), and pitch ($\theta_i$) commands for $N$ low-level attitude controllers in $N$ ARC-units. The yaw angle of each ARC-unit ($\psi_i^r$) is controlled by its low-level attitude controller.}\label{fig:control_diagram}
\end{figure}

Furthermore, let $N$ be the number of ARC-units, $L_i$ the link connecting unit $i$ to $i+1$, $J_{i,j}$ the joint connecting the ARC-unit $i$ to the link in the middle of ARC-unit $i$ and $j$, $d_{i,j} \in \mathbb{R}^3$ the vector from the Center-of-Gravity (CoG) of ARC-unit $i$ to the joint $J_{i,j}$ expressed in $\mathbb{B}_i$, and $l_i\in \mathbb{R}^3$ the vector from joint $J_{i+1,i}$ to joint $J_{i,i+1}$ expressed in $\mathbb{B}_{L_i}$ ($l_i\parallel {}^{L_i} e_1$, where ${}^{L_i} e_1$ is the unit vector in x-axis of $\mathbb{B}_{L_i}$ frame), $[x_i^r,y_i^r,z_i^r]^T$ the position of ARC-unit $i$ expressed in $\mathbb{W}$, ${R}_i \in \mathpzc{SO}(3)$ the rotation matrix from the $\mathbb{B}_i$ frame to $\mathbb{W}$, ${R}_{L_i}\in \mathpzc{SO}(3)$ the rotation matrix from $\mathbb{B}_{L_i}$ to $\mathbb{W}$. For $1\leq i \leq N-1$:

\small
\begin{equation}
\begin{split}
[x_{i+1}^r,y_{i+1}^r,z_{i+1}^r]^T = [x_i^r,y_i^r,z_i^r]^T - R_{i+1} d_{i+1,i} - R_{L_i} l_i + R_i d_{i,i+1} \label{eq:x_i} 
\end{split}
\end{equation}
\normalsize
Denote $[\phi_i^r,\theta_i^r,\psi_i^r]^T$ and $[\phi_i^l,\theta_i^l,\psi_i^l]^T$ respectively as roll, pitch, yaw Euler angles ($zyx$ order) of quadrotor $i$ and link $i$; $R_y,R_z$ the rotation matrices about the corresponding axes. We make two assumptions here: $(A1)$ the offsets between the joints and robots' COGs are negligible ($d_{i+1,i}\approx 0, d_{i,i+1} \approx 0$); $(A2)$ in normal operating conditions, tilt angles of every ARC-units are small. Also, $R_{L_i}l_i = R_z(\psi_i^l) R_y(\theta_i^l) l_i$ since $l_i\parallel {}^{L_i} e_1$. Hence for $1\leq i \leq N-1$, Eq.~(\ref{eq:x_i}) can be rewritten as:

\small
\begin{equation}
[x_{i+1}^r,y_{i+1}^r,z_{i+1}^r]^T = [x_i^r,y_i^r,z_i^r]^T - R_z(\psi_i^l) R_y(\theta_i^l) l_i  
\label{eq:x_i_approx} 
\end{equation}
\normalsize
From assumptions $(A1)$, $(A2)$ and Eq.~(\ref{eq:x_i_approx}), we choose the configuration space $\zeta$ of the ARC system as $[x_1^r,y_1^r,z_1^r,\psi_1^r,\\
\theta_1^l,\psi_1^l,\theta_2^l,\psi_2^l,\dots,\theta_{N-1}^l,\psi_{N-1}^l]^T$ which has dimension equal to $2N+2$. We utilize a control structure for the shape controller consisting of an MPC-based position controller for the head (first ARC-unit) as in~\cite{mpc_rosbookchapter} and $N-1$ parallel $\mathpzc{SO}(3)$ angular controllers for $N-1$ links based on~\cite{lee2010geometric}. The control diagram is shown in Figure~\ref{fig:control_diagram}.

%
%

\section{PROPOSED APPROACH}\label{sec:approach}
\begin{figure}[h!]
\centering
    \includegraphics[width=0.9\columnwidth]{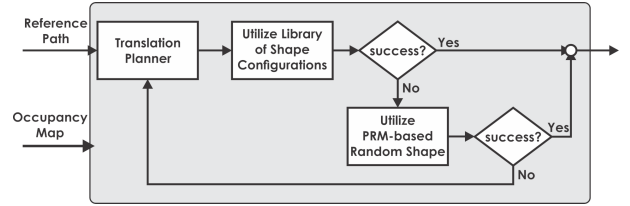}
\caption{Block diagram overview of the proposed Aerial Robotic Chain planner (ARCplanner).}\label{fig:arcplanner_architecture}
\vspace{-1ex}
\end{figure}

The proposed motion planning approach for the Aerial Robotic Chain integrating an arbitrary number of ARC-units, called ARCplanner, enables such a re-configurable aerial robotic system-of-systems to navigate complex environments in a computationally efficient manner. The ARCplanner employs a 2-step process of first attempting to find path segments that navigate from the current robot location to its goal, and subsequently identifying the best collision-free ARC shape configuration. For each path segment of the translation planning solution, the method first checks within a library of configurations which stores shape solutions if a collision-free solution based on one of the pre-considered configurations is possible. The library of ARC shape configurations contains two sets of solutions, namely a) optimized for the smallest possible cross-section, and b) optimized for maximum sensor coverage given a certain model of the frustum of the perception system onboard each ARC-unit. If a collision-free solution relying on the stored shape configurations is not found, the method proceeds to utilize a pre-sampled set of random configurations found through Probabilistic RoadMap (PRM) search of the configuration space for solutions regarding different ARC shapes in order to identify one that enables collision-free guidance for the specific path segment at hand. It is important to note that this process is invoked only sparingly and it is also critical to mention that the random configurations are sampled a priori and thus limiting the penalty on online computational cost. Finally, if and only if no shape configuration allows to navigate across the planned path segment, then the method re-samples solutions for the robot translation on the basis that different topologies at that level may facilitate collision-free navigation for its possible shape configurations. The overall architecture of the ARCplanner is depicted in Figure~\ref{fig:arcplanner_architecture}. 

%
%



\subsection{Translation Planner}\label{sec:transplanner}

The first step of the ARCplanner is a PRM-based motion planning process that is responsible for identifying collision-free paths for the leading robot of the Aerial Robotic Chain. In that sense, this planner searches for admissible configurations in the $\xi=[x_1^r,y_1^r,z_1^r]^T$ space (leading robot location) given a roadmap build phase, outlined in Algorithm~\ref{alg:transplanner}, and subsequent fast (multi-)query requests for paths. Function $\textrm{SampleFreeT}$ samples configurations $\xi_i$ sampled from a uniform distribution, while $\textrm{NearT}$ returns the other sampled vertices within a radius of $r_T$ from a vertex $v_T$. Function $\textrm{CollisionFreeT}$ evaluates if a connection between two sampled vertices $v_T,u_T$ (corresponding to two leading robot configurations) is collision-free given an occupancy map representation of the environment based on the work in~\cite{hornung13auro}. As Algorithm~\ref{alg:transplanner} builds the roadmap, subsequent multi-query collision-free paths can be derived seamlessly through lazy evaluation~\cite{choset2005principles} and thus for any feasible navigation problem a path solution $\sigma_T$ is derived from the graph $\mathbb{G}_T$. 

\begin{algorithm}[h]
\caption{Translation Planner}
\label{alg:transplanner}
\begin{algorithmic}[1]
\State $\mathbb{V}_T\gets~\xi_{0} \cup \{\textrm{SampleFreeT}(\xi_i)\}_{i=1,...,M_{\mathbb{V}_T}-1}; \mathbb{E}_T\gets~0$
\ForAll{$v_T \in \mathbb{V}_T$}
 \State $\mathbb{U}_T\gets~\textrm{NearT}(\mathbb{G}_T=(\mathbb{V}_T,\mathbb{E}_T),v_T,r_T) \setminus \{v_T\} ;$
 \ForAll{$u_T \in \mathbb{U}_T$}
    \If{$\textrm{CollisionFreeT}(v_T,u_T)$}
        \State $\mathbb{E}_T\gets \mathbb{E}_T\cup \{(v_T,u_T),(u_T,v_T)\};$ 
    \EndIf
 \EndFor
\EndFor
\State \Return $\mathbb{G}_T=(\mathbb{V}_T,\mathbb{E}_T),~\sigma_T;$
\end{algorithmic}
\end{algorithm}

\subsection{Library of Shape Configurations}\label{sec:lsc}

\begin{figure}[h!]
\centering
    \includegraphics[width=0.99\columnwidth]{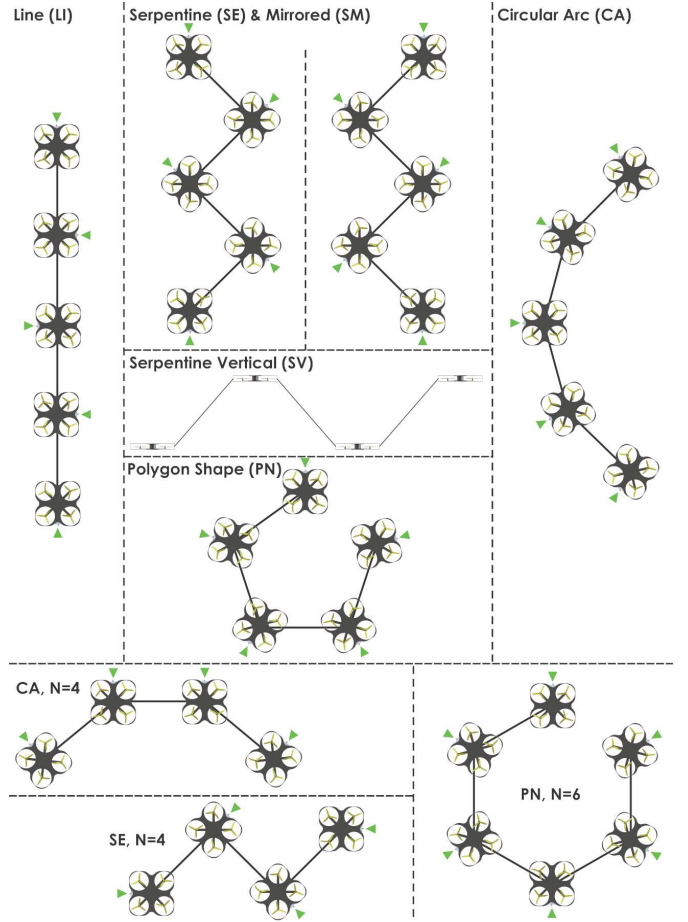}
\caption{Instances of the ARC shapes stored in the ``Library of ARC Shape Configurations'' (LSC). For the case of $N=5$ with the joint at the robot's Center of Gravity, the five possible configurations (LI, SE, SM, CA, PN) are presented, and indicative directions of the onboard frustum-constrained sensor of each robot are also depicted. For $N=4$ and the joint beyond the center of gravity, the SV configuration is presented. At the bottom of the figure, an arc shape (CA) for $N=4$ robots, a serpentine (SE) for $N=4$, and a polygon (PN) for $N=6$ are also presented. In our simulation studies, we have accounted for joint placements that enable the SV shape. }\label{fig:lacshapes}
\vspace{-1ex}
\end{figure}

As the dimensionality of the motion planning problem that needs to be solved to enable collision-free navigation is large for an ARC with a large number $N$ of ARC-units (e.g., $N \ge 4$) and since in most cases, certain geometric configurations tend to enable finding an admissible solution, a first step towards computational efficiency is to introduce a Library of ARC Shape Configurations (LSC). LSC consists of solutions optimized for a) narrow cross-section navigation, b) passing through corners, and c) maximizing the collective field-of-view of the robotic system-of-systems given an appropriately placed frustum-constrained sensor onboard each ARC-unit. As such, for every $N$ ARC-units, LSC may contain a) a ``Line'' configuration (LI), b) three serpentine configurations, specifically two (mirrored) on the x-y plane and one on x-z (SE, SM, SV), c) a Circular Arc (CA), and d) a Polygon Shape (PN). In the PN configuration for $N$ ARC-units, the robots are positioned by constructing the associated polygon through defining an inscribing circle and specifying $N$ equally-spaced nodes. In the CA configuration, the position of the robots is found by utilizing a $90^\circ$ arc and increasing/decreasing the radius as required to fit the ARC system given the length $l_r$ of the connecting rod. The (SE, SM, SV) configurations are defined based on the maximum angle of the joint which in turn defines the serpentine shape, while their orientations are aligned with the connecting rods. The LI configuration is straightforward. The shapes in LSC are stored in a list denoted $\mathbb{L}_{LSC}$. Finally, it is noted that for every configuration in LSC, a set of $n_{\psi}$ discrete yaw rotations $\{\psi_1^r\}$ of the head robot are considered during the collision-free path evaluation phase. Figure~\ref{fig:lacshapes} depicts instances of LSC configurations.

%
%

\subsection{Set of Random Shapes}\label{sec:srs}

The configurations stored in the LSC tend to provide admissible collision-free solutions most of the time and thus lead to a major reduction in computation time, while exploiting the shape-reconfiguration abilities of the ARC. However, it is not impossible that certain motion planning problems for the ARC navigating in complex and confined environments cannot be solved using only LSC shapes (or their rotations around the azimuth). To allow our approach to be able to provide a solution for such cases, a further sampling-based planning stage is considered. In particular, a PRM-based method for shape configuration sampling is implemented and aims to provide admissible solutions for paths sampled by the translation planner when no LSC-based solution is feasible. Considering a hovering configuration of the leading robot, say in the arbitrary location $\xi =[x_1^r,y_1^r,z_1^r]^T$, the PRM stage will sample for the leading robot and the ARC links and thus propose multiple shape configurations in an effort to identify new attainable solutions. Thus, the configuration space ($\eta$-Space) of this sampling-based planning stage is $\eta = [\psi_1^r,\theta_1^l, \psi_1^l, \theta_2^l, \psi_2^l, ...., \theta_{N-1}^l, \psi_{N-1}^l]^T$, where $\psi_1^r$ is the heading of the leading robot, $\theta_i^l,\psi_i^l$ are the pitch and yaw angle of the connecting link $i$, and $N$ is the number of ARC-units in the overall chain. It is highlighted that this step is executed offline and without any consideration of the specific map $\Ms$, while the $\eta$-space is chosen such that all shape transitions do not self-collide (straight edge in $\eta$-space by linearly varying the joint angles among configurations). Algorithm~\ref{alg:shapeplanner} outlines the main algorithmic steps of the random shape planer for the ARC system-of-systems based on PRM. This step is essential for solution resourcefulness and although in most cases LSC-configurations are sufficient, it was identified that in complex environments the ability to utilize new random shapes is key for successful and fast navigation. In the context of this algorithm, the function $\textrm{SampleFreeC}$ samples random $\eta$ shape configurations corresponding to $v_S$ vertices in the $\eta$-Space graph, and $\textrm{NearC}$ identifies a finite set of vertices that are within a radius $r_S$ from $v_S$. It is critical to highlight that this step is executed a priori, i.e., before the robot is even deployed to any environment, and as a result, a large number $N_{SRS}$ of randomly sampled shape configurations, a Set of Random Shapes (SRS) ($\mathbb{L}_{SRS}$), are stored in its memory ready to be utilized online when needed.  

\begin{algorithm}[h]
\caption{Random Shape Planner}
\label{alg:shapeplanner}
\begin{algorithmic}[1]
\State $\mathbb{V}_S\gets~\eta_{init} \cup \{\textrm{SampleFreeC}_i\}_{i=1,...,M-1}; \mathbb{E}_S\gets~0$
\ForAll{$v_S \in \mathbb{V}_S$}
 \State $\mathbb{U}_S\gets~\textrm{NearC}(\mathbb{G}=(\mathbb{V}_S,\mathbb{E}_S),v_S,r_S) \setminus \{v_S\} ;$
 \ForAll{$u_S \in \mathbb{U}_S$}
        \State $\mathbb{E}_S\gets \mathbb{E}_S\cup \{(v_S,u_S),(u_S,v_S)\};$ 
 \EndFor
\EndFor
\State \Return $\mathbb{G}_{SRS} \gets (\mathbb{V}_S,\mathbb{E}_S),~\mathbb{L}_{SRS};$
\end{algorithmic}
\end{algorithm}


\subsection{SRS-based Edge Connection}\label{sec:edgecon}

SRS provides the required resourcefulness of random shape configurations to facilitate transition between two points in complex environments. Let $\mathbb{L}_{SRS}$ represent the list of randomly sampled shape configurations in $\mathbb{G}_{SRS}$. The associated SRS-based edge connection algorithm is presented in Algorithm~\ref{alg:srsconnector}. In this context, a Valid Shape is one that does not collide with the environment at the considered starting vertex location of the leading robot $\xi_i$. Similarly, an Admissible Shape is one that remains collision free along the edge $(\overline{\xi_i,\xi_j})$ that connects two leading robot locations $\xi_i$ and $\xi_j$ ($v_T,u_T$ in the $\mathbb{G}_T$ graph). Furthermore, all Valid Shapes are kept as they may correspond to intermediate vertices in the $\eta$-Space allowing to reach the required Admissible Shapes. Thus, the graph $\mathbb{G}_{Val}$ is constructed from all valid shapes with a radius of connection $d_c$ in the $\eta$-Space. The function $\textrm{OptAdmissibleTransition}$ identifies the path corresponding to the minimum distance to an admissible shape in the configuration space, while similarly the method $\textrm{CollisionFreeE}$ checks for the whole edge $\sigma_{\min}$ to be collision-free. For all the cases that such an edge is not collision-free, the graph $\mathbb{G}_{Val}$ is pruned. Finally, the method returns $\sigma_{\min}$ starting from $v_C$ configuration in $\zeta=[x_1^r,y_1^r,z_1^r,\psi_1^r,\theta_1^l, \psi_1^l, \theta_2^l, \psi_2^2, ...., \theta_{N-1}^l, \psi_{N-1}^l]^T$ and traversing along $v_T,u_T$ ($(\overline{\xi_i,\xi_j})$), alongside a flag $s_v^u$ indicating planning success. 

\begin{algorithm}[h]
\caption{$\textrm{SRSConnect}(v_S,(v_T,u_T))$}
\label{alg:srsconnector}
\begin{algorithmic}[1]
\State $\mathbb{L}_{SRS}^{Val} \gets \textrm{ValidShapes}(\mathbb{L}_{SRS},\xi_i)$
\State $\mathbb{L}_{SRS}^{Adm} \gets \textrm{AdmissibleShapes}(\mathbb{L}_{SRS}^{Val},(\overline{\xi_i,\xi_j}))$
\State $\mathbb{G}_{Val} \gets \textrm{ConstructGraph}(\mathbb{L}_{SRS}^{Val},d_c)$
\State $b_{CF} \gets \textrm{FALSE}$ 
\While{$b_{CF} = \textrm{FALSE}$}
    \State $\sigma_{\min}(v_T,u_T) \gets \textrm{OptAdmissibleTransition}(\mathbb{G}_{Val})$
    \State $b_{CF} \gets \textrm{CollisionFreeE}(\sigma_{\min}(v_T,u_T))$
    \If{$b_{CF}=\textrm{FALSE}$}
        \State $\mathbb{G}_{Val} \gets \small \textrm{PruneCollidingEdges}\normalsize(\mathbb{G}_{Val},\sigma_{\min}(v_T,u_T))$
    \EndIf
\EndWhile

\State \Return $\{\sigma_{\min}(v_T,u_T),s_v^u\}$
\end{algorithmic}
\end{algorithm}

\subsection{Constructing the Solution Online}\label{sec:consol}

Considering the library of shape configurations (LSC) and $N_{SRS}$ a priori randomly sampled new ARC shapes in SRS, both stored in the memory of the robot, alongside the online identified leading robot path $\sigma_T$ connecting to the desired location $\xi_{goal}$ without the head robot colliding, the algorithm is now ready to proceed to the final steps of identifying an optimized solution for the full ARC. More specifically, in a first step considering all edges $(\overline{\xi_i,\xi_j})$ in $\sigma_T$ (connecting $(v_T,u_T)$) in the $\mathbb{G}_{T}$ graph, the method searches inside LSC to identify such shapes that provide collision-free navigation across $\sigma_T$. For that purpose, let $u_S$ denote each vertex in the shape $\eta$-Space corresponding to a location of the leading robot and thus to a vertex in $\mathbb{G}_{T}$, and $u_C$ the respective vertex in the $\zeta=[x_1^r,y_1^r,z_1^r,\psi_1^r,\theta_1^l, \psi_1^l, \theta_2^l, \psi_2^2, ...., \theta_{N-1}^l, \psi_{n-1}^l]^T$ space. Then, the solutions inside LSC are searched in the following manner: a) from the current robot shape first check if without any change of shape, a certain edge is admissible (i.e., collision-free for the full ARC consisting of $N$ ARC-units), b) if this is not possible consider discrete $\{\psi_1^r\}$ azimuth rotations of that shape about the leading robot, c) if still no solution is found then search within LSC in such an order such that shapes that are faster to be achieved are evaluated first. For a solution to be admissible, the shape must not be in collision at the beginning of a certain edge, at the end, and during the path. Although LSC configurations tend to provide solutions in most cases, it is still possible that certain path edges remain inadmissible with any LSC shape. When such a condition appears, the method proceeds to evaluate if any of the configurations in SRS provides a solution for the current edge. In this process, solution admissibility is considered in the same sense as when evaluating LSC configurations. The overall steps are presented in Algorithm~\ref{alg:soleval}, while when no solution is found for certain edges, the flag $\{s_v^u\}$ returns false and thus the translation planner is re-invoked (a rare case) for the remaining path to the destination vertex $v_{goal}$ (with associated lead robot location $\xi_{goal}$). Furthermore, the method $\textrm{MotionPlan}$ reconstructs the path $\sigma_{ARC}$ in the full configuration space $\zeta$ from the edges $\mathbb{E}_{sol}$. Identification of the optimal path is based on minimizing the following cost-to-transition $C_T$:

\vspace{-2ex}
\small
\begin{eqnarray}
&C_T = \biggl\{(N)(\psi_{1,c_1}^r-\psi_{1,c_2}^r)^2 + \nonumber \\
  &\sum_{i=1}^{N-1}(N-i)[(\psi_{i,c_1}^l - \psi_{i,c_2}^l)^2 + (\theta_{i,c_1}^l-\theta_{i,c_2}^l)^2]\biggr\}^{\!1/2}
\end{eqnarray}
\normalsize
where $N$ is the number of ARC-units, $c_1,c_2$ are the initial and final shape configurations and $\psi_i^l,\theta_i^l$ correspond to the yaw and pitch angles of link $i$. It is noted that, during a transition between two vertices in the $\zeta$-Space, the robot first makes the shape transition ($\eta$-space) and then the translation transition ($\xi$-space).

\begin{algorithm}[h]
\caption{Online Solution Construction}
\label{alg:soleval}
\begin{algorithmic}[1]
\State $\mathbb{E}_{sol}\gets~0$
\ForAll{edges $(v_T,u_T) \in \sigma_T$}
    \State $v_S\gets$ starting vertex of current edge;
    \State $b_v^u = \textrm{FALSE};$
    \ForAll{$\{\psi_1^r\}$ rotations}
        \ForAll{$\eta_{lsc} \in \mathbb{L}_{LSC}$}
            \State $u_S \gets$ vertex for rotated $\eta_{lsc};$ 
            \If{$\textrm{Admissible}(v_S,u_S,v_T,u_T)$}
                \State $\mathbb{E}_{sol}\gets \mathbb{E}_{sol}\cup \{(v_C,u_C)\};$ 
                \State $b_v^u = \textrm{TRUE};$
            \EndIf
        \EndFor
    \EndFor
    \If{$b_v^u = \textrm{FALSE}$}
        \State $s_v^u = \textrm{TRUE};$
        \State $\tiny\{\sigma_{\min}(v_T,u_T),s_v^u\} \gets \textrm{SRSConnect}(v_S,(v_T,u_T))\normalsize$
        \If{$s_v^u = \textrm{FALSE}$}
            \State $\sigma_T \gets \textrm{TranslationPlannerQuery}(v_T, v_{goal})$
        \Else
            \State $\mathbb{E}_{sol}\gets \mathbb{E}_{sol}\cup \{\sigma_{\min}(v_T,u_T)\};$
        \EndIf
    \EndIf
\EndFor
\State $\sigma_{ARC} = \textrm{MotionPlan}(\mathbb{E}_{sol});$
\State \Return $\sigma_{ARC};$ 
\end{algorithmic}
\end{algorithm}

\subsection{Computational Complexity}

The proposed algorithm is designed so as to provide computationally efficient solutions for the motion planning problem of complex multi-linked aerial robotic chains with multiple ARC-units. Therefore it relies on the a priori stored LSC and SRS. Then online, it further prioritizes attempts to find solutions through shapes that tend to satisfy most common environments stored in LSC, while SRS configurations are considered only when needed. As such, the steps that contribute online computational cost are the translation planner (Section~\ref{sec:transplanner}), the SRS-based edge connection (only when invoked) (Section~\ref{sec:edgecon}), and the online solution construction (Section~\ref{sec:consol}). Focusing on the most important cost factors, given an ARC-unit size modeled as a cube with length $D_R$, then, for the translation planner $\textrm{SampleFreeT}$ is $\BigO(V_M/V_{M,free} \times V_{D_R}/r_v^3 \times \log(V_M/r_v^3))$. Similarly, for the $\textrm{CollisionFreeT}$ function, the cost is $\BigO(V_{D_R}/r_v^3 \times d_{avg}/r_v \times \log(V_{M}/r_v^3))$, where $d_{avg}$ is the average edge length. Similarly, for the step of online construction of the solution and regardless if an LSC or SRS configuration is considered, the computational cost for the $\textrm{Admissible}$ function given an average edge length $d_{avg}$ and $N$ ARC-units is $\BigO(N\times V_{D_R}^\prime/r_v^3 \times d_{avg}^C/r_v \times \log(V_{M}/r_v^3))$. A key question in terms of worst-case analysis is the number of shape configurations $N_{SRS}$ within SRS. Although the majority of cases are not searched, the absolute worst-case is to scale the previously mentioned computational cost term by $N_{SRS}$. Lastly, during the SRS-based connection to build $\mathbb{G}_{Val}$, a similar computational cost term holds for $N-1$ robots (as at this point the head robot is not considered). This analysis emphasizes only on the dominant cost terms and thus it is partial though indicative of the factors that dominate in scaling the complexity of the problem given an Octomap-based occupancy-map~\cite{hornung13auro} representation.

\section{EVALUATION STUDIES}\label{sec:eval}
In order to comprehensively evaluate the proposed motion planning algorithm for the Aerial Robotic Chain, four sets of simulation studies were conducted using the Gazebo-based RotorS Simulator~\cite{furrer2016rotors}. First, we present two sets of studies that relate to the application of the ARCplanner in two room-like environments, one of which involved a narrow window. In these studies we do not only present the end result of the ARCplanner but also evaluate the contribution of its components. A third study compares the ARCplanner with a full-state PRM thus demonstrating that given the high dimensionality of the underlying motion planning problem a direct application of PRM over the full state is not practical. Last, we present an application-driven demonstration of our approach. Recorded videos of these results are available at \url{https://youtu.be/wP8Mg9_YEh8}.

\subsection{Environment 1: Room with Window}

\begin{figure}[h!]
\centering
    \includegraphics[width=0.99\columnwidth]{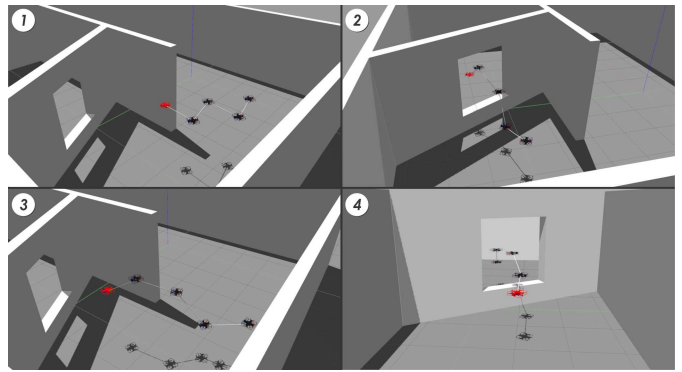}
\caption{Instances of the Gazebo-based simulation of an ARC with $N=5$ ARC-units navigating a room with a window based on the ARCplanner. }\label{fig:env1gazebo}
\vspace{-1ex}
\end{figure}

\begin{figure}[h!]
\centering
    \includegraphics[width=0.99\columnwidth]{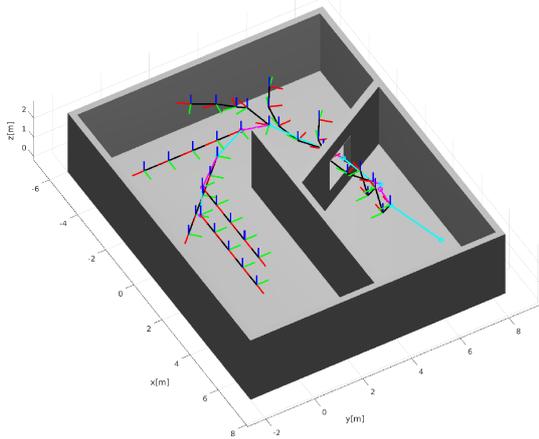}
\caption{ARCplanner solution for an ARC with $N=5$ ARC-units navigating a room with a window. The planner makes $9$ changes in shape. Each robot is depicted with a coordinate frame, while alternating cyan and magenta paths are used between planning steps. }\label{fig:env1result}
\end{figure}

In a first simulation study, two cases of ARC systems, namely with $N=4$ and $N=5$ ARC-units were commanded to navigate a room involving a narrow window and sub-spaces divided with a wall (``Environment 1''). The relevant simulation result for the case of $N=5$ robots in the chain are presented in Figures~\ref{fig:env1gazebo} and ~\ref{fig:env1result}. As depicted, the ARCplanner successfully commands the robot to change its shape and pass through the most narrow settings. 

%
%

%
%

To allow us to study in detail the role of LSC and SRS in the solution resourcefulness of ARCplaner we conduct a specific study where ARCplanner either utilizes only LSC, only SRS or both. As this environment is structured in a manner that it has one narrow window whereas the rest is relatively free, we observe from the experiments that it is far easier for the robot to evaluate and decide on a shape from LSC when near relatively free regions. While this works in rather open regions, when ARC plans a path obliquely to a given window, it becomes fairly difficult for it to be able to translate using the limited LSC shapes. In fact, it is found that the use of SRS is crucial for finding solutions in such challenging cases. However, as discussed in Section~\ref{sec:approach}, the process of finding solutions through SRS is expensive (Section~\ref{sec:edgecon}) since the planner spends significant time for a solution that can be arrived at easily, hence increasing the planning time. Thus as the conducted simulations show, it is the overall combined use of LSC and SRS that offers both the necessary planning resourcefulness and the computational efficiency for a planner capable of finding solutions in challenging environments within reasonable time. Table~\ref{tab:env1} summarizes the findings of this study indicating how the performance increases in case of combined LSC/SRS use versus each case alone. Within Table~\ref{tab:env1}, $M_{\mathbb{V}_T}$ denotes the number of vertices in the translation planning $\mathbb{G}_T$ graph,  $N$ is the number of ARC-units in the considered ARC robot, $t_P$ is the overall planning time, and $t_{\mathbb{G}_T}=25.6s$ is the computational time required to build this graph. The ``LSC'' column denotes the ARCplanner using only LSC and not SRS, the ``SRS'' column corresponds to the mirrored case, while ``Both'' stands for the full ARCplanner utilizing LSC and SRS.

\begin{table}[h!]
\caption{Key Statistics for Environment 1}
\label{tab:env1}
\begin{center}
\begin{tabular}{|l|l|l|l|
>{\columncolor[HTML]{EFEFEF}}l |
>{\columncolor[HTML]{EFEFEF}}l |
>{\columncolor[HTML]{EFEFEF}}l |}
\hline
                   & \textbf{LSC} & \textbf{SRS} & \textbf{Both} & \textbf{LSC} & \textbf{SRC} & \textbf{Both} \\ \hline
$N$                & $4$          & $4$          & $4$           & $5$          & $5$          & $5$           \\ \hline
$M_{\mathbb{V}_T}$ & $1500$       & $1500$       & $1500$        & $1500$       & $1500$       & $1500$        \\ \hline
$N_{SRS}$ & $0$       & $500$       & $500$        & $0$       & $500$       & $500$        \\ \hline
$t_P$              & $11.3$       & $23.0$       & $4.4$         & $46.1$       & $54.4$       & $4.4$         \\ \hline
\end{tabular}
\end{center}
\end{table}

\subsection{Environment 2: Maze-like Room}

\begin{figure}[h!]
\centering
    \includegraphics[width=0.99\columnwidth]{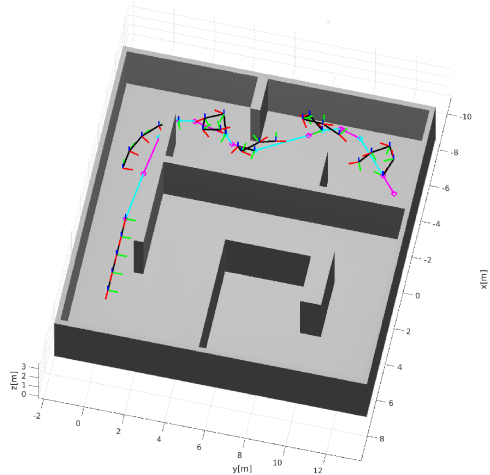}
\caption{ARCplanner solution for an ARC with $N=5$ robots navigating a maze-like room. The planner makes in total $10$ changes in shape.}\label{fig:env2result}
\vspace{-1ex}
\end{figure}

In a second study in a maze-like room (Environment 2), the robot has to follow a zig-zag path. Being constrained in this manner, the robot has to find an appropriate shape-change path that allows it to cross such complicated surroundings. In this environment, since not enough space is available for the rear part of the robot to move freely, the LSC shapes which are mostly oriented linearly fail to provide solutions. Therefore, the planner uses SRS to find an appropriate solution. A relevant simulation study with $N=5$ ARC-units is depicted in Figure~\ref{fig:env2result}. Furthermore, a comparative study analogous to the one conducted in the previous case when only using LSC, only SRS, or the full ARCplanner is summarized in Table~\ref{tab:env2}. In this case, the time to build graph $\mathbb{G}_T$ is $t_{\mathbb{G}_T}=9.4s$.

%
%

\begin{table}[h!]
\caption{Key Statistics for Environment 2}
\label{tab:env2}
\begin{center}
\begin{tabular}{|l|l|l|l|
>{\columncolor[HTML]{EFEFEF}}l |
>{\columncolor[HTML]{EFEFEF}}l |
>{\columncolor[HTML]{EFEFEF}}l |}
\hline
                   & \textbf{LSC} & \textbf{SRS} & \textbf{Both} & \textbf{LSC} & \textbf{SRC} & \textbf{Both} \\ \hline
$N$                & $5$          & $5$          & $5$           & $6$          & $6$          & $6$           \\ \hline
$M_{\mathbb{V}_T}$ & $1500$       & $1500$       & $1500$        & $1500$       & $1500$       & $1500$        \\ \hline
$N_{SRS}$ & $0$       & $500$       & $500$        & $0$       & $500$       & $500$        \\ \hline
$t_P$              & $1.5$       & $9.5$       & $1.27$         & $32.7$       & $14.2$       & $7.9$         \\ \hline
\end{tabular}
\end{center}
\end{table}

\subsection{Comparison against full-state PRM}

An alternative to the proposed approach would be to utilize a full-state PRM, utilizing Lazy evaluation~\cite{choset2005principles}, within which an online PRM would sample for the leading robot position and all the pitch and yaw Euler angles of the links in an effort to find collision-free paths for ARC. A relevant study was conducted based on Environment 1. The results are summarized in Table~\ref{tab:comparefullstate}, where $M_F$ stands for the vertices sampled in full-state PRM, $t_G$ denotes the time for building the graphs needed in the different cases, and $t_P$ the time spent for planning. As depicted, when the full-state PRM uses a number of vertices sampled leading to a time to build the graph and time to plan similar to that of the ARCplanner, it most often fails ($80\%$ of the time) to find a solution, while when we tune the number of vertices to safely find solutions it ends up requiring orders of magnitude greater computational cost. This result is aligned with the theoretical expectation for the cost of full-state PRM for multi-DoF systems.

\begin{table}[]
\caption{Comparison against full-state PRM}
\label{tab:comparefullstate}
\begin{center}
\begin{tabular}{|l|l|l|
>{\columncolor[HTML]{EFEFEF}}l |}
\hline
                                                           & \multicolumn{2}{l|}{\textbf{Full-State PRM}} & \textbf{ARCplanner}                                                              \\ \hline
\begin{tabular}[c]{@{}l@{}}Planning\\ Density\end{tabular} & $M_F=1500$               & $M_F=15000$               & \begin{tabular}[c]{@{}l@{}}$M_{\mathbb{V}_T}=1500$,\\ $N_{SRS}=500$\end{tabular} \\ \hline
$t_G$                                                      & $48.1$               & $3712.8$              & $26.5$                                                                           \\ \hline
$t_P$                                                      & $20.8$             & $127.3$             & $15.4$                                                                       \\ \hline
Success $\%$                                               & $20\%$               & $100\%$               & $100\%$                                                                          \\ \hline
\end{tabular}
\end{center}
\end{table}

\subsection{Application-driven Demonstration}

\begin{figure}[h!]
\centering
    \includegraphics[width=0.86\columnwidth]{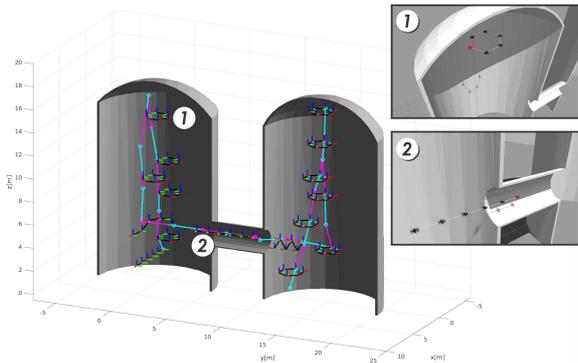}
\caption{ARCplanner solution for $N=6$ ARC-units navigating two-silo structure and optimizing, when possible, for sensor coverage at any instance of the robot.}\label{fig:silodemo}
\vspace{-1ex}
\end{figure}

Finally, we present an application-driven simulation study where an ARC-unit is commanded to navigate among certain waypoints in a two-silo structure. The robot is commanded to optimize sensing coverage, thus prescribing the use of the Polygon Shape (``PN'' configuration) when possible, and only employ shapes that minimize its cross-section when going through constrained regions (e.g., between the two silos). Figure~\ref{fig:silodemo} depicts the relevant simulation result. 

%
%

\section{CONCLUSIONS}\label{sec:concl}

An approach for the motion planning problem of multilinked aerial robotic chains consisting of quadrotor Micro Aerial Vehicles was presented. The method utilizes an engineered library of shape configurations, as well as shapes identified through a probabilistic roadmap and facilitates efficient navigation despite the high-dimensionality of the underlying problem. The efficiency of the proposed approach is presented within multiple simulations and is compared against a standard full-state PRM solution.

\bibliography{main}             

\end{document}